# Color Image Clustering using Block Truncation Algorithm

Dr. Sanjay Silakari [1], Dr. Mahesh Motwani [2] and Manish Maheshwari [3]

[1] **Department of Computer Science and Engineering**
Rajiv Gandhi Technical University
Bhopal, M.P., India
*ssilakari@yahoo.com*

[2] **Department of Computer Science and Engineering**
Jabalpur Engineering College
Jabalpur, M.P., India
*mahesh_9@hotmail.com*

[3] **Department of Computer Science & Application**
Makhanlal C. National University of Journalism and Communication
Bhopal, M.P., India
*manishbhom@yahoo.com*

**Abstract**
With the advancement in image capturing device, the image data been generated at high volume. If images are analyzed properly, they can reveal useful information to the human users. Content based image retrieval address the problem of retrieving images relevant to the user needs from image databases on the basis of low-level visual features that can be derived from the images. Grouping images into meaningful categories to reveal useful information is a challenging and important problem. Clustering is a data mining technique to group a set of unsupervised data based on the conceptual clustering principal: maximizing the intraclass similarity and minimizing the interclass similarity. Proposed framework focuses on color as feature. Color Moment and Block Truncation Coding (BTC) are used to extract features for image dataset. Experimental study using K-Means clustering algorithm is conducted to group the image dataset into various clusters.

***Key words:*** *Image features, Clustering, Color moments, BTC*

## 1. Introduction

The rapid progress in computer technology for multimedia system has led to a rapid increase in the use of digital images. Rich information is hidden in this data collection that is potentially useful in a wide range of applications like Crime Prevention, Military, Home Entertainment, Education, Cultural Heritage, Geographical Information System (GIS), Remote sensing, Medical diagnosis, and World Wide Web [1, 2]. Rich information is hidden in these data collection that is potentially useful. A major challenge with these fields is how to make use of this useful information effectively and efficiently. Exploring and analyzing the vast volume of image data is becoming increasingly difficult.

The image database containing raw image data cannot be directly used for retrieval. Raw image data need to be processed and descriptions based on the properties that are inherent in the images themselves are generated. These inherited properties of the images stored in feature database which is used for retrieval and grouping. The strategy for earlier image retrieval system focused on "search-by-query". The user provides an example image for the query, for which the database is searched exhaustively for images that are most similar.
Clustering is a method of grouping data objects into different groups, such that similar data objects belong to the same group and dissimilar data objects to different clusters [3,4]. Image clustering consists of two steps the former is feature extraction and second part is grouping. For each image in a database, a feature vector capturing certain essential properties of the image is computed and stored in a feature base. Clustering algorithm is applied over this extracted feature to form the group.
In this paper we propose a data mining approach to cluster the images based on color feature. Concept of color moment is extended to obtain the features and k_means algorithm is applied to cluster the images. The rest of paper is organized as follows: In section two we provide overview of the previous work related to image retrieval and mining. Section three introduces the concept of 3.1 Color moments, 3.2 Block Truncation Coding Algorithm, 3.3 K-means clustering algorithm. In section four we

IJCSI

present the results of our experiments and finally section five concludes the paper.

## 2. Related Work

Feature extraction is the process of interacting with images and performs extraction of meaningful information of images with descriptions based on properties that are inherent in the images themselves. Color information is the most intensively used feature for image retrieval because of its strong correlation with the underlying image objects. Color Histogram [2] [5] [6] is the commonly and very popular color feature used in many image retrieval system. The mathematical foundation and color distribution of images can be characterized by color moments [7]. Color Coherence Vectors (CCV) have been proposed to incorporate spatial information into color histogram representation [8]. Texture refers to the presence of a spatial pattern that has some properties of homogeneity. Textures are replications, symmetries and combinations of various basic patterns or local functions, usually with some random variation. There are a number of texture features which have been used frequently liked Tamura Texture feature [5,6], Simultaneous Auto-Regressive (SAR) models[9], Gabor texture features[10] and Wavelet transform features[11,12].

Intelligently classifying image by content is an important way to mine valuable information from large image collection. [13] Explore the challenges in image grouping into semantically meaningful categories based on low-level visual features. The concept of fuzzy ID3 decision tree for image retrieval was discussed in [14]. ID3 is a decision tree method based on Shannon's information theory. Given a sample data set described by a set of attributes and an outcome, ID3 produces a decision tree, which can classify the outcome value based on the values of the given attributes like Color, Texture and Spatial Location. Image dataset were defined in 10 classes (concepts): grass, forest, sky, sea, sand, firework, sunset, flower, tiger and fur. At each level of the ID3 decision tree, the attribute with smallest entropy is selected from those attributes not yet used as the most significant for decision-making.

The SemQuery [15] approach proposes a general framework to support content-based image retrieval based on the combination of clustering and querying of the heterogeneous features. Given a query image, the SemQuery compares the features of query image to those of database images, resulting in a group of retrieved image sets based on individual feature classes. Database images were categorized into five categories of cloud, floral, leaves, mountain and water. Hierarchical clustering approach was performed on texture and color feature. Wavelet transforms extract the texture features while color histograms method used for color feature. [16] Describe data mining and statistical analysis of the collections of remotely sensed image. Large images are partitioned into a number of smaller more manageable image tiles. Then those individual image tiles are processed to extract the feature vectors.

## 3. Proposed Work

An image is a spatial representation of an object and represented by a matrix of intensity value. It is sampled at points known as pixels and represented by color intensity in RGB color model. A basic color image could be described as three layered image with each layer as Red, Green and Blue as shown in fig 1.

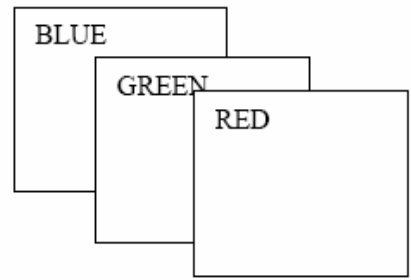

Fig. 1 Image Components

### 3.1 Color Feature Extraction

Color moments are measures that can be used differentiate images based on their features of color. The basis of color moments lays in the assumption that the distribution of color in an image can be interpreted as a probability distribution. Probability distributions are characterized by a number of unique moments (e.g. Normal distributions are differentiated by their mean and variance). It therefore follows that if the color in an image follows a certain probability distribution, the moments of that distribution can then be used as features to identify that image based on color. Stricker and Orengo [7] use three central moments of an image's color distribution in which $p^k_{ij}$ is the value of the k-th color component of the ij-image pixel and P is the height of the image, and Q is the width of the image. They are Mean, Standard deviation and Skewness.

MOMENT 1 – Mean:

$$E_k = \frac{1}{PQ} \sum_{i=1}^{P} \sum_{j=1}^{Q} p^k_{ij} \qquad (1)$$

Mean can be understood as the average color value in the image.

**MOMENT 2 - Standard Deviation:**

The standard deviation is the square root of the variance of the distribution.

$$SD_k = \sqrt{\frac{1}{PQ} \sum_{i=1}^{P} \sum_{j=1}^{Q} (p^k_{ij} - E_k)^2} \qquad (2)$$

**MOMENT 3 – Skewness:**

$$S_k = \left( \frac{1}{PQ} \sum_{i=1}^{P} \sum_{j=1}^{Q} (p^k_{ij} - E_k)^3 \right)^{1/3} \qquad (3)$$

Skewness can be understood as a measure of the degree of asymmetry in the distribution.

### 3.2 Block Truncation Algorithm

Steps in Block Truncation Coding Algorithm:

1. Split the image into Red, Green, Blue Components

2. Find the average of each component
    Average of Red component
    Average of Green component
    Average of Blue component

3. Split every component image to obtain RH, RL, GH, GL, BH and BL images
RH is obtained by taking only red component of all pixels in the image which are above red average and RL is obtained by taking only red component of all pixels in the image which are below red average. Similarly GH, GL, BH and BL can be obtained.

4. Apply color moments to each splitted component i.e. RH, RL, GH, GL, BH and BL.

5. Apply clustering algorithm to find the clusters.

### 3.3 K Means Clustering Algorithm

K-means is one of the simplest unsupervised learning algorithms in which each point is assigned to only one particular cluster. The procedure follows a simple, easy and iterative way to classify a given data set through a certain number of clusters (assume k clusters) fixed a priori. The procedure consists of the following steps,

Step 1: Set the number of cluster k

Step 2: Determine the centroid coordinate

Step 3: Determine the distance of each object to the centroids

Step 4: Group the object based on minimum distance

Step 5: Continue from step 2, until convergence that is no object move from one group to another.

## 4. Experiments

The proposed scheme has been performed using a image database of 1000 images including 10 classes, which can be downloading from the website http://**wang.ist.psu.edu/iwang/test1.tar.** Each class has 100 images. Each image is of size 384*256 pixels. The system is developed in Matlab. We define unichrome feature as values that are extracted from a single color layer of Red, Green and Blue.

Original Image (Bus Image)

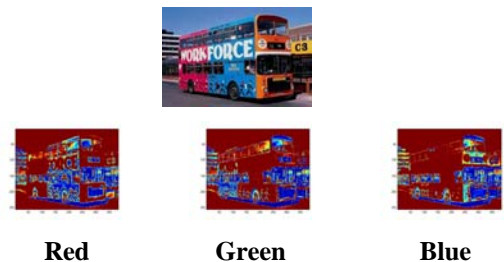

**Red**      **Green**      **Blue**

Fig. 2 RGB Components of Sample Image

The first part of evaluation computes color moments for each of the three color components. Each color component yields a feature vector of three elements as discussed in section 3.1 i.e. mean, standard deviation and skewness. Thus total nine feature vectors are calculated for one image.

In the second part apply Block Truncation Coding Algorithm discussed in section 3.2 over RH, RL, GH, GL, BH and BL. Thus total 18 feature vectors are calculated for one image.

Table 1: Recall and Precision using Color Moments

| Classes | Recall | Precision |
|---|---|---|
| African People and villages | 30 | 25.43 |
| Beaches | 47 | 40.17 |
| Buildings | 33 | 23.57 |
| Buses | 37 | 35.24 |
| Dinosaurs | 100 | 92.59 |
| Elephants | 32 | 35.56 |

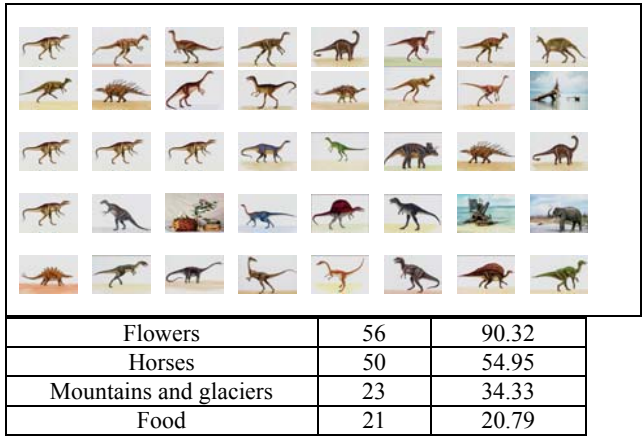

| Flowers | 56 | 90.32 |
|---|---|---|
| Horses | 50 | 54.95 |
| Mountains and glaciers | 23 | 34.33 |
| Food | 21 | 20.79 |

Fig. 3 Sample Dinosaurs Clusters using color moments

Based on commonly used performance measures in information retrieval, two statistical measures were computed to assess system performance namely Recall and Precision.
Recall consists of the proportion of target images that have been retrieved among all the relevant images in the database.

Recall  =  Number of Relevant Images Retrieved
                Total Number of Relevant Images            (4)

Precision consists of the proportion of relevant images that are retrieved.

Precision =  Number of Relevant Images Retrieved
                      Total Retrieved Images                    (5)

Table 2: Recall and Precision using BTC

| Classes | Recall | Precision |
|---|---|---|
| African People and villages | 44 | 33.58 |
| Beaches | 42 | 42.42 |
| Buildings | 8 | 7.92 |
| Buses | 52 | 44.83 |
| Dinosaurs | 99 | 97.06 |
| Elephants | 39 | 44.83 |
| Flowers | 80 | 94.12 |
| Horses | 50 | 58.82 |
| Mountains and glaciers | 35 | 43.21 |
| Food | 25 | 22.32 |

Figure 3 showing the sample images in Dinosaurs cluster using color moments and Figure 4 showing the sample images in Flowers cluster using BTC algorithm.

Table 1 and Table 2 shows the values of recall and precision of each classes using color moments and BTC algorithms. Precision is maximized for dinosaur's images in both the clustering algorithm. For maximum classes recall and precision of BTC is better than Color moment feature extraction.

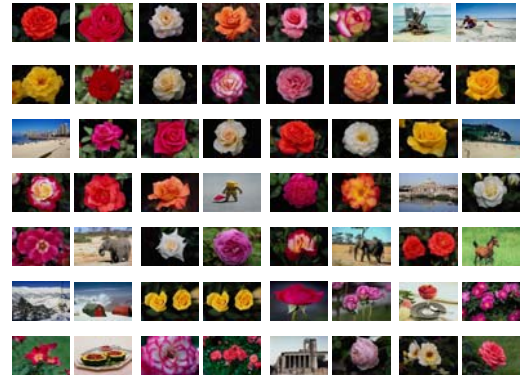

Fig. 4 Sample Flowers Clusters using BTC

## 5. Conclusion

In image retrieval system, the content of an image can be expressed in terms of different features such as color, texture and shape. These low-level features are extracted directly from digital representations of the image and do not necessarily match the human perception of visual semantics. We proposed a framework of unsupervised clustering of images based on the color feature of image. Test has been performed on the feature database of color moments and BTC. K-means clustering algorithm is applied over the extracted dataset. Results are quite acceptable and showing that performance of BTC algorithm is better than color moments.

IJCSI International Journal of Computer Science Issues, Vol. 4, No. 2, 2009                                                                    35[6]  Alex Pentland, Rosalind W. Picard, Stan Sclaroff, "Photobook: Tools for Content-Based Manipulation of Image Databases", Storage and Retrieval for Image and Video Databases (SPIE) 1994: 34-47

[7]  M.Stricker and M.Orengo, "Similarity of color images", Storage and Retrieval for Image and Video Databases III (SPIE) 1995: 381-392

[8]  Greg Pass, Ramin Zabih, Justin Miller: Comparing Images Using Color Coherence Vectors. ACM Multimedia 1996: 65-73

[9]  Jianchang Mao, Anil K. Jain, "Texture classification and segmentation using multiresolution simultaneous autoregressive models", Pattern Recognition 25(2): 173-188 (1992)

[10] B. S. Manjunath, Wei-Ying Ma, "Texture Features for Browsing and Retrieval of Image Data", IEEE Trans. Pattern Anal. Mach. Intell. 18(8): 837-842 (1996)

[11] Ingrid Daubechies, "The wavelet transform, time-frequency localization and signal analysis", IEEE Transactions on Information Theory 36(5): 961-1005 (1990)

[12] Tianhorng Chang, C. C. Jay Kuo, "Texture analysis and classification with tree-structured wavelet transform", IEEE Transactions on Image Processing 2(4): 429-441 (1993)

[13] Y. Uehara, S. Endo, S. Shiitani, D. Masumoto, and S. Nagata, "A computer-aided Visual Exploration System for Knowledge Discovery from Images", In Proceedings of the Second International Workshop on Multimedia Data Mining (MDM/KDD'2001), San Francisco, CA, USA, August, 2001.

[14] Ying Liu1, Dengsheng Zhang1, Guojun Lu1 , Wei-Ying Ma2, "Deriving High-Level Concepts Using Fuzzy-Id3 Decision Tree for Image Retrieval", IEEE 2005, pp 501-504

[15] Gholamhosein Sheikholeslami, Wendy Chang, Aidong Zhang, "SemQuery: Semantic Clustering and Querying on Heterogeneous Features for Visual Data", IEEE Trans. Knowl. Data Eng. 14(5): 988-1002 (2002)

[16] Krzysztof Koperski, Giovanni Marchisio, Selim Aksoy, and Carsten Tusk, "Applications of Terrain and Sensor Data Fusion in Image Mining", IEEE 2002, pp 1026-1028

[17] H.B.Kekre, Tasneem Mirza, "Content Based Image Retrieval using BTC with Local Average Threshholding". In Proceeding of ICCBIR 2008, pp 5-9

[18] http://wang.ist.psu.edu/iwang/test1.tar
IJCSI